\DocumentMetadata{
	lang		= eng-US, 	
	pdfstandard	= A-3u,		
	pdfversion	= 1.7, 		
}

\documentclass[colorlinks,upint,subscriptcorrection,varvw,hyphenate,balance,nohead,nofoot]{asmeconf} 

\usepackage{multirow}
\usepackage{xcolor}
\definecolor{lightyellow}{RGB}{255,255,180}
\usepackage[normalem]{ulem}
\usepackage{amsmath}
\usepackage{xcolor}
\DeclareMathOperator*{\argmax}{arg\,max}

\hypersetup{%
	pdfauthor={Fatemeh Elhambakhsh{,} Gaurav Ameta{,} Aditi Roy{,} Hyunwoong Ko},
	pdftitle={Boundary Representation-Driven Graph Transformer Modeling for Machining Operation Sequencing},
	pdfkeywords={Machining Process Planning, Boundary Representation, B-Rep, Graph Transformers},
	pdfsubject={Machining operation sequence prediction from B-Rep geometry},
}

\allowdisplaybreaks 

\begin{document}

%

\title{Boundary Representation-Driven Graph Transformer Modeling for Machining Operation Sequencing} 
 
%
%
%

\SetAuthors{%
    Fatemeh Elhambakhsh\affil{1}, 
	Gaurav Ameta\affil{2}, 
	Aditi Roy\affil{2},  
	Hyunwoong Ko\affil{3}\CorrespondingAuthor{Hyunwoong.Ko@asu.edu}
	}

\SetAffiliation{1}{School of Computing and Augmented Intelligence, Arizona State University, Tempe, AZ 85281, USA}
\SetAffiliation{2}{Siemens Foundational Technologies,
Princeton, NJ 08540, USA}
\SetAffiliation{3}{School of Manufacturing Systems and Networks, Arizona State University, Mesa, AZ 85212, USA}


\maketitle





\begin{abstract}
Machining process planning is inherently complex due to the interdependence between part geometry, machining operations, and evolving manufacturable states. Traditional approaches rely heavily on expert knowledge and rule-based systems, limiting scalability and automation. While recent machine learning methods have shown promise, most operate on simplified geometric representations, such as voxels or hand-crafted features, which fail to fully exploit the rich geometric and topological information encoded in Boundary Representation (B-Rep) models.
In this work, we propose a graph transformer–based framework for geometry-to-sequence learning that directly predicts machining operation sequences from B-Rep design geometry. The proposed method constructs topology-aware graphs from B-Rep entities, where faces and their adjacencies are encoded as node and edge features capturing both geometric and relational information. A graph neural network encoder learns geometry-aware embeddings, which are then used to condition a dual-decoder transformer that autoregressively generates machining operation sequences. This formulation enables the model to capture both operation-type prediction and process-level ordering constraints directly from design geometry, without requiring intermediate feature extraction or machining-state representations.
Experiments on a dataset of 2,991 synthetic machined parts demonstrate strong performance, achieving 95\% accuracy for main operation prediction, 64\% accuracy for sub-operation prediction, and 93\% accuracy for sequence length prediction. Further analysis shows that the model effectively captures precedence relationships and geometry-dependent constraints inherent in machining processes. These results highlight the potential of B-Rep–driven graph learning for scalable and automated machining process planning.
\end{abstract}

\section{Introduction}

Boundary representation (B-Rep) is a foundational modeling scheme in computer-aided design (CAD) that defines a solid object by explicitly describing its enclosing boundary surfaces, curves, points, and other topological constructs \cite{zou2025bringing}. A B-Rep model organizes a solid into a hierarchy of topological entities: a body consists of one or more shells, each shell is composed of faces bounded by loops, loops are made of oriented edges, and edges connect at vertices, with each topological entity linked to corresponding geometric primitives such as surfaces, curves, and points \cite{lee2001partial}. This structure encodes both geometry and topology, enabling precise modeling of shape and connectivity, including face adjacencies and edge sharing, which are essential for evaluating geometric and topological quantities, such as manifold type and genus. The comprehensive integration of geometry and topology in B-Rep models underpins their use in modern CAD systems, enabling downstream applications such as simulation and automated process planning \cite{elhambakhsh2025mp}.
Yet, despite B-Rep's representational richness, automatically extracting manufacturing-relevant knowledge—such as machining operation sequences—directly from B-Rep geometry leads to ambiguity and remains unsolved.

Machining process planning (MP) is a systematic activity that converts raw materials into a finished part through a sequence of machining operations, such as milling, turning, and drilling \cite{afif2025computer}. 
MP determines the order of operations, defines tool paths, selects cutting tools, and sets parameters \cite{afif2025computer}. It serves as a bridge between CAD \cite{elhambakhsh2025domain} and Manufacturing, enabling  CAD-based complex product geometries to be translated into Computer Aided Machining (CAM)-based machining steps \cite{wang2024machining}, ensuring production efficiency, productivity, and final part quality \cite{marzia2023automated}. Traditional MP heavily relies on expert knowledge, which constrains efficient decision-making due to limited automation capabilities \cite{besharati2019intelligent}. 

Machine learning (ML) techniques have now emerged as enablers of MP automation, learning process patterns from data and reducing the need for manual decision-making \cite{elhambakhsh2025mp, elhambakhsh2025amdiffusion}. In particular, recent ML studies in MP have leveraged advanced methods such as graph neural networks (GNNs) and large language models to model complex machining processes and generate process knowledge \cite{wang2024machining, xu2025large, sabri2020monitoring, elhambakhsh2021developing, elhambakhsh2022scan, elhambakhsh2023latent}, thereby shifting toward automated planning. Despite these advances, most existing ML-based process planning methods operate on 
simplified geometric representations such as voxels or hand-crafted feature graphs, 
failing to fully exploit the rich geometric and topological information encoded in B-Rep 
models~\cite{maqueda2025deepms}. This limits their ability to capture fine-grained surface relationships 
and adjacency structures that are critical for accurate operation sequencing. 

To address this gap, we propose a graph transformer-based
framework that constructs topology-aware graphs directly from
B-Rep entities and predicts machining operation sequences
directly from design geometry without requiring intermediate
machining-state representations or geometric simplification.
The framework encodes face-level geometric descriptors and
topological adjacencies through a GNN encoder and generates
ordered machining operation sequences autoregressively through
transformer decoders conditioned on the learned part embedding.
Unlike prior voxel-based approaches that lose precise geometric
information, feature-recognition methods that require intermediate
manufacturing abstractions, and prior frameworks that rely on
intermediate machining-state representations, the proposed
approach directly learns geometry-to-sequence relationships
from topology-aware B-Rep graph representations of original design
geometry. Specifically, this work contributes: (1) a B-Rep graph
formulation that preserves exact geometric and topological
relationships between faces and edges, (2) a direct geometry-to-
sequence learning framework that eliminates intermediate feature
extraction and dynamic machining-state modeling, and (3) a
dual-label autoregressive prediction framework that jointly predicts
both primary operations and fine-grained sub-operations directly
from design geometry.

The remainder of this paper is organized as follows: Section~\ref{sec:litreview} reviews 
related work, Section~\ref{sec:method} describes the proposed methodology, 
Section~\ref{sec:casestudy} presents the experimental results and analysis, and 
Section~\ref{sec:conclusion} concludes the paper.

\section{Literature Review}
\label{sec:litreview}
The application of ML to manufacturing process planning from CAD geometry represents a convergence of two critical research streams: learning effective representations of B-Rep models and automating MP through ML. We organize the related work into these two main categories.

\subsection{B-Rep Representation Learning}

The fundamental challenge in applying ML to B-Rep models is how to represent the geometric-topological structure complexities of B-Rep models. BRepNet~\cite{lambourne2021brepnet} pioneered neural processing by defining convolutional kernels over oriented coedges, using topological walks (next, previous, mate relationships) for message passing directly on the B-Rep manifold structure. UV-Net~\cite{jayaraman2021uv} took an alternative approach by sampling regular grids from the parametric UV domains of B-Rep surfaces and curves, coupling image convolutional neural networks for local geometry with graph convolutional networks for global topology.

Hierarchical CADNet~\cite{colligan2022hierarchical} proposed a two-level hierarchical representation: a face adjacency graph encoding topological relationships between faces and mesh patch graphs encoding fine-grained surface geometry. BRepGAT~\cite{lee2023brepgat} represented B-Rep as face-level graphs with attention mechanisms to dynamically weight neighboring face importance, defining feature descriptors capturing surface type, curvature, and adjacency. BRepFormer~\cite{dai2025brepformer} introduced a pure transformer-based approach that encodes B-Rep structure through edge-feature and topology-feature biases integrated into attention computation. AAGNet~\cite{wu2024aagnet} introduced geometric attributed adjacency graphs that preserve topological, geometric, and extended attributes from neutral B-Rep files, enabling multi-task learning across semantic segmentation, instance segmentation, and bottom face identification.

\subsection{Machining Process Planning}

Traditional MP depends heavily on expert knowledge and rule-based systems, which hinder efficient automation and scalability~\cite{besharati2019intelligent}. ML approaches have recently emerged to enable process planning automation by learning manufacturing patterns directly from data, thereby reducing reliance on manual expertise and decision-making. Some studies focused on machining feature recognition as a bridge between CAD geometry and process planning. Yeo et al.~\cite{yeo2021machining} extracted B-Rep face-level descriptors for classifying machining feature types. Peddireddy et al.~\cite{peddireddy2021identifying} demonstrated that 3D CNNs with transfer learning could classify machining processes and assess manufacturability from voxelized CAD models. While these methods successfully identify Manufacturing process planning features, they do not address the critical problem of operation sequencing.

In this regard, Zhang et al.~\cite{zhang2024novel} introduced deep reinforcement learning (DRL) for machining process route planning, representing parts as directed attributed adjacency graphs processed by convolutional GNNs, with an actor-critic architecture treating planning as a Markov decision process. Wu et al.~\cite{wu2021fast} extended DRL to simultaneous operation sequencing and resource selection under dynamic conditions, using masking algorithms to enforce precedence constraints. However, both DRL approaches require explicit feature graphs as input rather than learning directly from raw geometry, necessitating a separate feature recognition stage.

Regarding geometry-to-sequence prediction, DeepMS~\cite{maqueda2025deepms} combines 3D CNN autoencoders with transformer decoders to predict ordered machining operation sequences directly from part geometry, trained on 1.08 million synthetic prismatic parts. However, it operates on voxelized representations at fixed resolution, potentially losing the precise geometric and topological information that BRep provides—such as exact surface types, edge curvatures, and topological adjacency—which are critical for manufacturing decisions. Brep2Seq~\cite{zhang2024brep2seq} demonstrated sequence-to-sequence learning from B-Rep using Graphormer architectures, though focused on CAD reconstruction rather than machining.

Elhambakhsh et al.~\cite{elhambakhsh2025mp} proposed MP-GFormer to predict 
machining operation sequences by learning intermediate machining and design 
geometries using a graph transformer approach. However, this method still relies 
on intermediate machining geometry representations, which require additional 
geometric processing beyond the raw design model. Our work addresses this 
limitation by proposing a transformer-based B-Rep graph learning framework that 
predicts machining operation sequences directly from design geometry, without 
requiring intermediate machining geometries, feature extraction, or any additional 
geometric processing at each machining step.

\section{Methodology}
\label{sec:method}
\begin{figure*}[!th]

\centering\includegraphics[width=\linewidth]{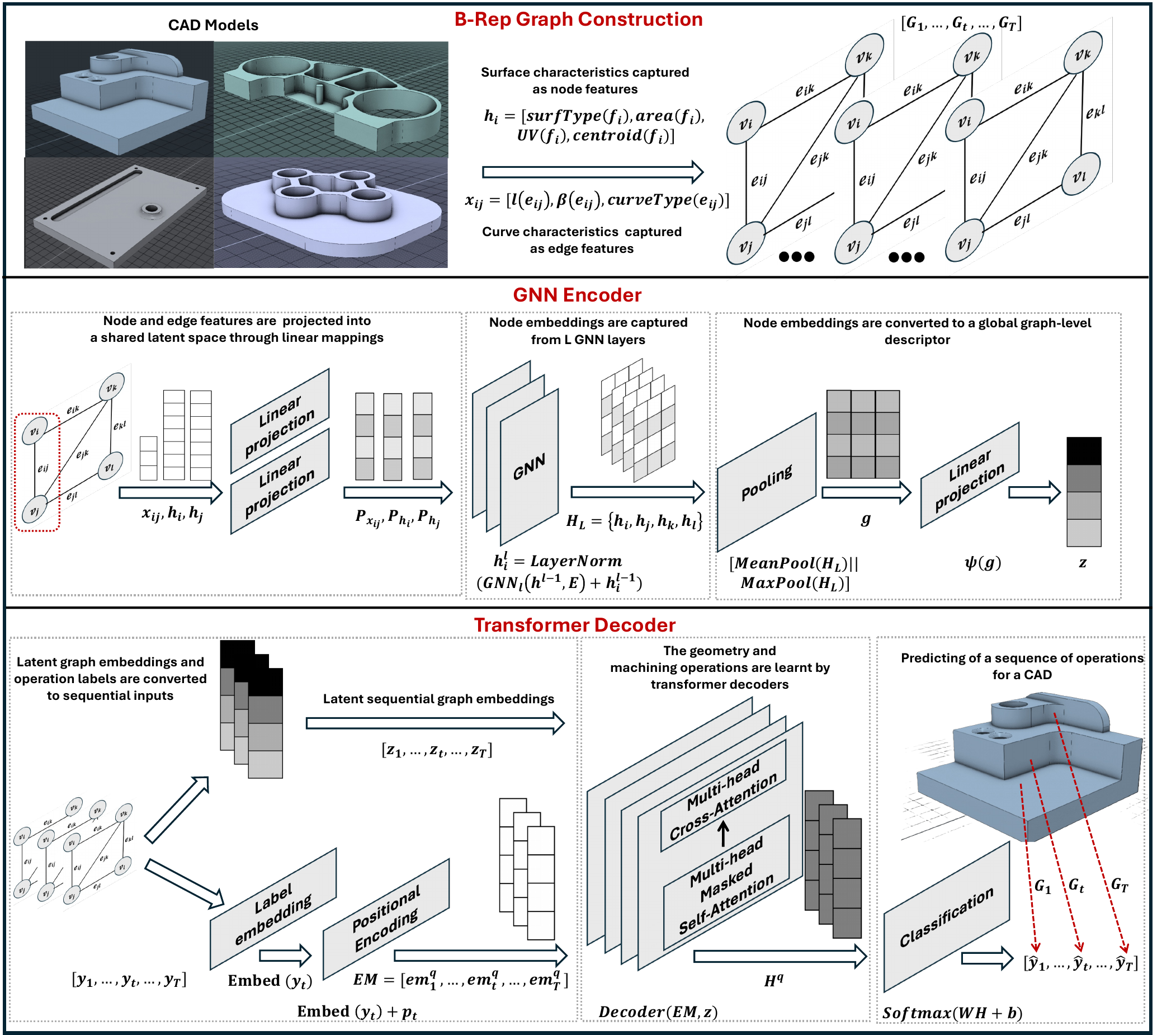}
\caption{Overview of the proposed method. CAD models are converted to graphs by capturing surface and curve features from B-Rep. The graph is fed into a GNN encoder to learn the dependencies between faces and edges of the part geometry. The learned features are then passed to a transformer decoder to autoregressively predict a sequence of machining operations directly from the design geometry graph.}\label{fig:1}
\end{figure*}
\subsection{Problem Formulation}
The design geometry of a machined part is represented as a B-Rep, which encodes precise geometric surfaces and their topological adjacencies.
To enable structured learning, the B-Rep is converted into a graph, $G$, where nodes and shared edges define relational surface connectivity.
This graph formulation preserves both local geometric attributes and inter-surface relationships, providing a structured representation of the part geometry that supports the learning of geometry-dependent manufacturing semantics.
Then, we formulate a geometry-to-sequence learning problem where, given the B-Rep of a part's design geometry $G$, our objective is to predict the ordered sequence of machining operations $\mathcal{S} = \{(y_1^{(1)}, y_1^{(2)}), (y_2^{(1)}, y_2^{(2)}), \ldots, (y_T^{(1)}, y_T^{(2)})\}$ required to manufacture the part. Here, $y_t^{(1)}$ denotes the machining operation type and $y_t^{(2)}$ represents a sub-operation type at manufacturing step $t$.

We aim to learn a mapping function $o: G \rightarrow \mathcal{S}$. 
This mapping is governed by feature-level geometric dependencies in $G$ and process-level precedence constraints among operations within $\mathcal{S}$.
Material removal operations typically follow structured transitions
(e.g., roughing preceding finishing),
and the feasibility of a given operation depends on
previous geometric transformations of the part.
Accordingly, the objective is formulated as learning the conditional distribution $\mathbb{P}(\mathcal{S} \mid G)$, where the sequential structure reflects intrinsic manufacturing process dependencies rather than arbitrary label ordering.
This formulation enables the model to capture both operation-type identification and process-level ordering logic directly from geometric input. The mapping function $o: G \rightarrow \mathcal{S}$ maximizes the likelihood as Equation \ref{o}:
\begin{equation}
\label{o}
o^* = \argmax_{o} \mathbb{P}(\mathcal{S} | G; \theta) = \argmax_{o} \prod_{t=1}^{T} \mathbb{P}(y_t^{(1)}, y_t^{(2)} | G, y_{<t}^{(1)}, y_{<t}^{(2)}; \theta)
\end{equation}
, where $\theta$ represents the learnable parameters and $y_{<t}$ denotes all operations preceding timestep $t$. 
For this purpose, we utilized: (1) A topology-aware GNN encoder for B-Rep, and (2) A transformer architecture for multi-label operations sequence generation.
The overall framework is illustrated in Figure.~\ref{fig:1}.

\subsection{B-Rep Graph Construction}

We construct geometric-topological graphs directly from the B-Rep models that represent the design geometry of parts to be machined. Figure \ref{fig:2} illustrates the data representation and graph construction pipeline.
Each part is converted into a graph $
G = (V, E)$, where nodes \(V\) correspond to B-Rep faces and edges \(E\) represent topological adjacencies between faces. Each face \(f_i \in V\) is treated as a node with a surface-level geometric descriptor vector as Equation \ref{eq:nodefeatures}.  
\begin{equation}
h^{\text{B-Rep}}_{i} =
\big[
\text{surfType}(f_i),\;
\text{area}(f_i),\;
\text{UV}(f_i),\;
\text{centroid}(f_i)
\big]
\in \mathbb{R}^{d_{\text{B-Rep}}}.
\label{eq:nodefeatures}
\end{equation}
, where \(\text{UV}(f_i)\) denotes the parametric surface coordinates, and \(\text{centroid}(f_i)\) captures the spatial location of the face. Edges represent face adjacencies. For each adjacent face pair \((f_i,f_j)\), we extract a compact relational descriptor as Equation \ref{eq:edgefeatures}.  
\begin{equation}
x^{\text{B-Rep}}_{ij} =
\big[
\ell(e_{ij}),\;
\beta(e_{ij}),\;
\text{curveType}(e_{ij})
\big]
\in \mathbb{R}^{3},
\label{eq:edgefeatures}
\end{equation}
where  
\(\ell(e_{ij})\) is the shared-edge length,  
\(\beta(e_{ij})\) is the dihedral angle, and  
\(\text{curveType}(e_{ij})\) denotes the geometric type of the connecting edge.

Topological adjacency relationships were extracted directly from
the B-Rep kernel by identifying pairs of faces that share a common
boundary edge. Each adjacent face pair $(f_i, f_j)$ forms an edge
in graph $G=(V,E)$, preserving exact topological connectivity
defined in the original CAD model. Categorical geometric
attributes such as surface type (e.g., planar, cylindrical, conical)
and curve type were encoded using one-hot representations,
while continuous geometric attributes including face area,
centroid coordinates, shared-edge length, and dihedral angles
were normalized prior to graph construction. The UV representation
was obtained from the parametric domain boundaries of each face,
capturing local surface parameterization information. Since
different CAD models contain varying numbers of faces and
adjacency relationships, the resulting B-Rep graphs naturally
vary in size and structural complexity, preserving the geometric
diversity of the original part designs.

The constructed B-Rep graph preserves machining-relevant geometric cues through the explicit encoding of surface attributes $h_i^{\text{B-Rep}}$ and relational descriptors $x_{ij}^{\text{B-Rep}}$.
In particular, the dihedral angle component $\beta(e_{ij})$ encodes local concavity and convexity between adjacent faces, capturing recessed and protruding geometric transitions that are relevant to material removal processes (e.g., pocket-like regions).
Similarly, the surface type $\text{surfType}(f_i)$ identifies analytic geometries such as cylindrical faces, which are indicative of drilling or boring operations.
Coplanar or smoothly connected face clusters, implicitly characterized through adjacency relations in $E$ and shared geometric parameters in $h_i^{\text{B-Rep}}$, often correspond to planar milling regions.
By jointly modeling node-level geometric descriptors and edge-level topological relations, the graph $G = (V,E)$ retains the structured information necessary for learning manufacturable feature abstractions directly from design geometry.

\textbf{\begin{figure*}
\centering\includegraphics[width=\linewidth]{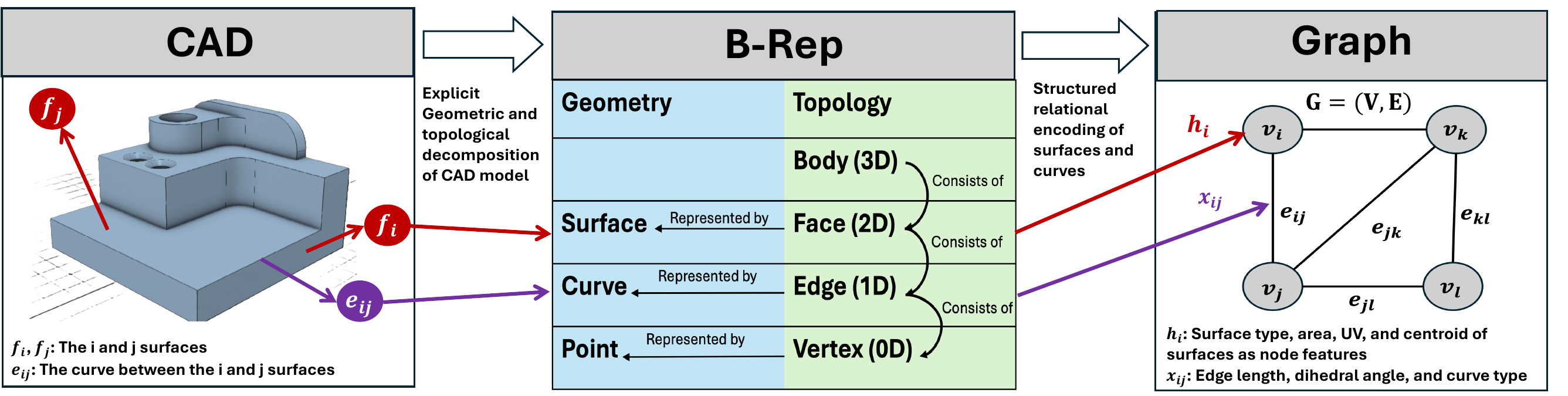}
\caption{Illustration of the representation pipeline. 
(a) CAD rendering with manufacturing-relevant faces highlighted. 
(b) Corresponding B-Rep view where faces $f_i$, $f_j$, and shared edges $e_{ij}$ are shown. 
(c) Constructed geometric–topological graph $G=(V,E)$: faces map to nodes with node attributes $h_i^{\text{B-Rep}}$, and adjacencies map to edges with relational descriptors $x_{ij}^{\text{B-Rep}}$. Arrows indicate the transformation steps from CAD geometry to B-Rep and then to the graph representation used by the encoder. The red and purple colors indicate the transformation of surfaces and their features as well as curves and their features from CAD to graph, respectively.}
\label{fig:2}
\end{figure*}}
Each B-Rep graph \(G\) corresponds to a sequence of machining operations with $T$ timesteps. Each timestep \(t\) includes two operations,
\(
y_t^{(1)}
\)
and
\(
y_t^{(2)}
\),
associated with two machining-operation taxonomies,  
namely the primary operation type and the secondary sub-operation category defined in the dataset.
This construction preserves both geometric richness and topological fidelity, enabling the learning framework to infer machining semantics directly from B-Rep design geometry.

\subsection{Operation Sequence Prediction}

Given a B-Rep graph representation \(G = (V, E)\) of a machined part, the goal is to predict the corresponding machining operation sequence. The framework maps each B-Rep graph to a fixed-dimensional embedding using a topology-aware GNN encoder, which then conditions a dual-decoder Transformer that generates two parallel operation-label sequences in an autoregressive manner.

\paragraph{B-Rep Graph Encoder.}
Node and edge features obtained from the B-Rep representation are first projected into a shared latent space through linear mappings with non-linear activation and layer normalization as $P_{h_{i}}$ and $P_{x_{ij}}$. 
This latent space provides a unified parameterization of the surface descriptors $h_i^{\text{BRep}}$ and relational attributes $x_{ij}^{\text{BRep}}$, 
embedding geometric quantities such as surface type, area, and dihedral angle into a continuous feature manifold.
By projecting heterogeneous B-Rep attributes into a common representation space, the model enables consistent aggregation of curvature, adjacency, and analytic surface information,
facilitating the propagation of machining-relevant geometric dependencies across the graph structure.
The encoder then applies a stack of alternating Graph Attention (GAT) and Graph Convolution (GCN) layers \cite{velivckovic2017graph, zhang2019graph}. For layer \(\ell\), the node update is given by Equation~\ref{gnn}:
\begin{equation}
h_i^{(\ell)} = 
\operatorname{LayerNorm}\!\Big(
\operatorname{GNN}_\ell(h^{(\ell-1)}, E) 
+ h_i^{(\ell-1)}
\Big)
\label{gnn}
\end{equation}, where even-numbered layers use multi-head attention and odd-numbered layers use convolution. 

The alternating GAT and GCN layers are designed to regulate how node features $h_i^{(\ell)}$ aggregate neighboring information derived from $x_{ij}^{\text{BRep}}$.
Since geometric interactions encoded by $\beta(e_{ij})$ and shared-edge connectivity in $E$ do not contribute uniformly to downstream machining decisions, GAT layers introduce attention coefficients that adaptively weight adjacent faces based on their geometric relevance.
This mechanism allows concave or curvature-discontinuous transitions to exert a stronger influence on feature updates when informative.
In contrast, GCN layers perform isotropic neighborhood aggregation, ensuring stable propagation of surface descriptors such as $\text{surfType}(f_i)$ and preserving local geometric consistency across smoothly connected regions.
By alternating adaptive and uniform aggregation, the encoder progressively integrates localized geometric relations into higher-order structural representations of the part geometry.

After $L$ layers of graph propagation, the node embeddings $H_L = \{h_i^{(L)}\}_{i \in V}$ encode localized geometric–topological information across the part.
To generate a fixed-dimensional representation suitable for sequence conditioning, these node-level features must be aggregated into a single graph-level descriptor.
Global mean and max pooling produce a graph-level feature:
\begin{equation}
g = 
\big[
\operatorname{MeanPool}(H_L) 
\;\|\;
\operatorname{MaxPool}(H_L)
\big],
\end{equation}
which is passed through a projection network to obtain the B-Rep embedding:
\begin{equation}
z = \psi(g) \in \mathbb{R}^{d}.
\end{equation}

Mean pooling computes $\frac{1}{|V|} \sum_{i \in V} h_i^{(L)}$, aggregating the final-layer node embeddings into a global descriptor that reflects the overall distribution of geometric–topological features across the part.
This operation preserves averaged contributions of surface attributes and adjacency-informed representations encoded in $H_L$.
Max pooling is defined as the element-wise maximum 
$\max_{i \in V} h_i^{(L)}$, which selects the most strongly activated 
geometric responses, emphasizing structurally dominant regions 
whose feature magnitudes significantly influence manufacturability.
The concatenation $[\mathrm{MeanPool}(H_L) \,\|\, \mathrm{MaxPool}(H_L)]$ therefore integrates global structural statistics with salient geometric signals into the final part-level embedding $g$.
The resulting embedding $z \in \mathbb{R}^d$ provides a fixed-dimensional geometric context vector for downstream sequence modeling.
\paragraph{Operation Embedding and Positional Encoding.}
To enable conditional generation of machining operations, the structured operation labels are reformulated as tokenized sequences compatible with Transformer-based autoregressive decoding.
Each machining sequence is expressed as two parallel token sequences, one for each operation taxonomy.  
For each operation label type \(q \in \{1,2\}\), the sequence is:
\[
y^{(q)} = 
(\langle s \rangle, y_1^{(q)},\dots,y_T^{(q)},\langle /s \rangle)
\]
, where \(\langle s \rangle\) and \(\langle /s \rangle\) denote START and END tokens.  
Tokens are embedded and augmented with sinusoidal positional encodings, as shown in Equation \ref{pt}.
\begin{equation}
em^{(q)}_t = \operatorname{Embed}(y^{(q)}_t) + p_t
\label{pt}
\end{equation}
The positional encoding $p_t$ injects permutation-sensitive structure into the otherwise order-invariant token embeddings, enabling the model to represent precedence-dependent operation sequences.
Stacking token embeddings yields 
$EM^{(q)} = [em^{(q)}_1, \dots, em^{(q)}_T] \in \mathbb{R}^{T \times d}$,
which serves as the decoder input sequence.

\paragraph{Transformer Decoders.}
The B-Rep embedding \(z\) serves as a shared geometric memory that conditions two parallel Transformer decoders, one for each operation taxonomy.
For operation label type $k$, the decoder maps the embedded sequence $EM^{(q)}$ and geometric context $z$ to hidden states as Equation \ref{decoder} \cite{vaswani2017attention}:
\begin{equation}
H^{(q)} = 
\operatorname{Decoder}^{(k)}(
EM^{(q)},\; z )
\label{decoder}
\end{equation}
, where masked self-attention models precedence-dependent operation transitions within taxonomy $q$, and cross-attention injects the geometry-conditioned embedding $z$ into the sequence states.
This separation enables taxonomy-specific evolution of operation tokens under a shared geometric constraint, ensuring that sequence generation remains consistent with the part-level geometric–topological representation.
The output logits at timestep $t$ are given by Equation \ref{softmax}:
\begin{equation}
\hat{y}^{(q)}_t = 
\operatorname{Softmax}\!\big(
W^{(q)} H^{(q)}_t + b^{(q)}
\big)
\label{softmax}
\end{equation}
, producing a categorical distribution over the discrete machining operation vocabulary associated with taxonomy $k$, thereby parameterizing the conditional probability $p_\theta(y_t^{(q)} \mid y_{<t}^{(q)}, z)$.

\paragraph{Training Objective.}
Model parameters $\theta$ are optimized via maximum likelihood estimation
under teacher forcing, where ground-truth tokens are provided as inputs to condition subsequent predictions during optimization.
For each label type \(q\), the conditional sequence likelihood factorizes autoregressively as Equation \ref{loss}:
\begin{equation}
\mathcal{L}^{(q)} =
- \sum_t 
\log p_\theta\!\left(
y_t^{(q)} \mid y_{<t}^{(q)}, z
\right)
\label{loss}
\end{equation}.
This objective maximizes the log-likelihood of each operation token given its preceding operations and the geometric conditioning vector, thereby enforcing temporal consistency within the sequence.
By optimizing token-level conditional probabilities, the model learns to generate operation sequences that satisfy temporal ordering constraints among operations while remaining consistent with geometry-dependent feasibility.
The overall objective aggregates the losses from both operation taxonomies as Equation \ref{TLoss}.
\begin{equation}
\mathcal{L} = 
\mathcal{L}^{(1)} + \mathcal{L}^{(2)}
\label{TLoss}
\end{equation}

\paragraph{Autoregressive Inference.}
During inference, each operation sequence is generated by recursively conditioning on previously predicted tokens and the geometric embedding $z$.
For operation label type $q \in \{1,2\}$, the prediction at timestep $t$ is given by Equation \ref{inf}:
\begin{equation}
\hat{y}^{(q)}_t =
\arg\max_c\; 
p_\theta(c \mid \hat{y}^{(q)}_{<t}, z)
\label{inf}
\end{equation}
, which terminates when the END token is emitted.
Both decoders remain conditioned on the shared embedding $z$, yielding taxonomy-specific sequences under a common geometric context.

\section{Case Study}
\label{sec:casestudy}

\subsection{Data Collection}

\begin{figure*}
\centering\includegraphics[width=\linewidth]{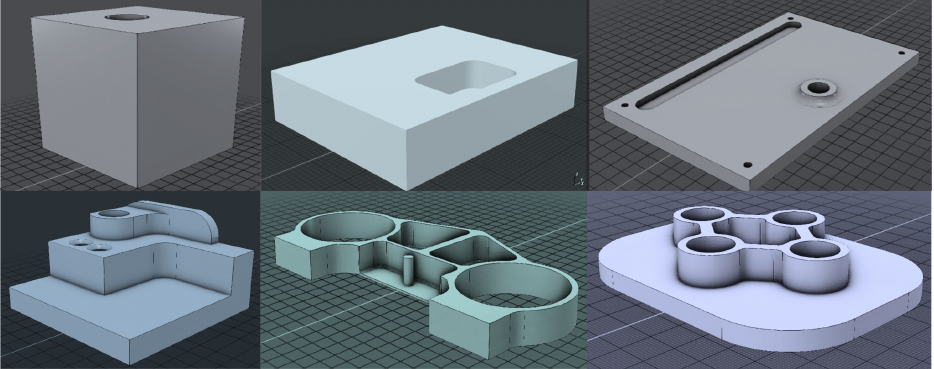}
\caption{Parts geometries used for synthesized data generation: (1) simple geometry
(top row) and (2) complex geometry (bottom row). The contrast between these two
classes is intentional, enabling evaluation of model generalization across varying
levels of topological and geometric complexity.}\label{Sample}
\end{figure*}

The dataset was generated using Siemens NX CAD/CAM and contains two file types:
(1) B-Rep files providing comprehensive topology and geometry of part designs, and
(2) text files containing machining operations and process information. Synthetic
data is organized into two classes: (1) simple and (2) complex geometry, as shown
in Figure~\ref{Sample}. Simple geometries consist of prismatic parts with a limited
number of features, such as single pockets or holes, while complex geometries involve
multiple interacting features, varying depths, and more intricate topological
structures. This deliberate contrast between geometry types is central to
evaluating the model's capacity to generalize across varying levels of manufacturing
complexity.


Each part geometry was parameterized by (a) overall dimensions (length, width,
height) and (b) feature specifications (pocket/hole locations and sizes). By
systematically varying parameter values across a defined range and automatically
updating the corresponding process plans through automated NX scripts, we generated
2,991 valid part-process pairs. Each pair consists of a B-Rep graph extracted from
the CAD model and a structured text file recording the associated machining operation
sequence, including operation types, sub-operation types, and their ordering. The
automated generation pipeline ensures consistency between part geometry and process
plan annotations, eliminating manual labeling errors and enabling controlled
evaluation of how geometry complexity influences predicted operation sequences.
Additionally, the observed class imbalance in Figures \ref{fig:labels} and \ref{fig:labels1} reflects authentic manufacturing practice, where foundational operations such as floor-wall milling and drilling are ubiquitous, while specialized operations such as countersinking and curve-drive machining are geometry-specific. This distribution aligns with
practical manufacturing workflows, where certain operations serve as foundational
building blocks for most parts while others address specialized functional
requirements.


\subsection{Experiments}

\begin{figure*}[tb!]
\centering\includegraphics[width=\linewidth]{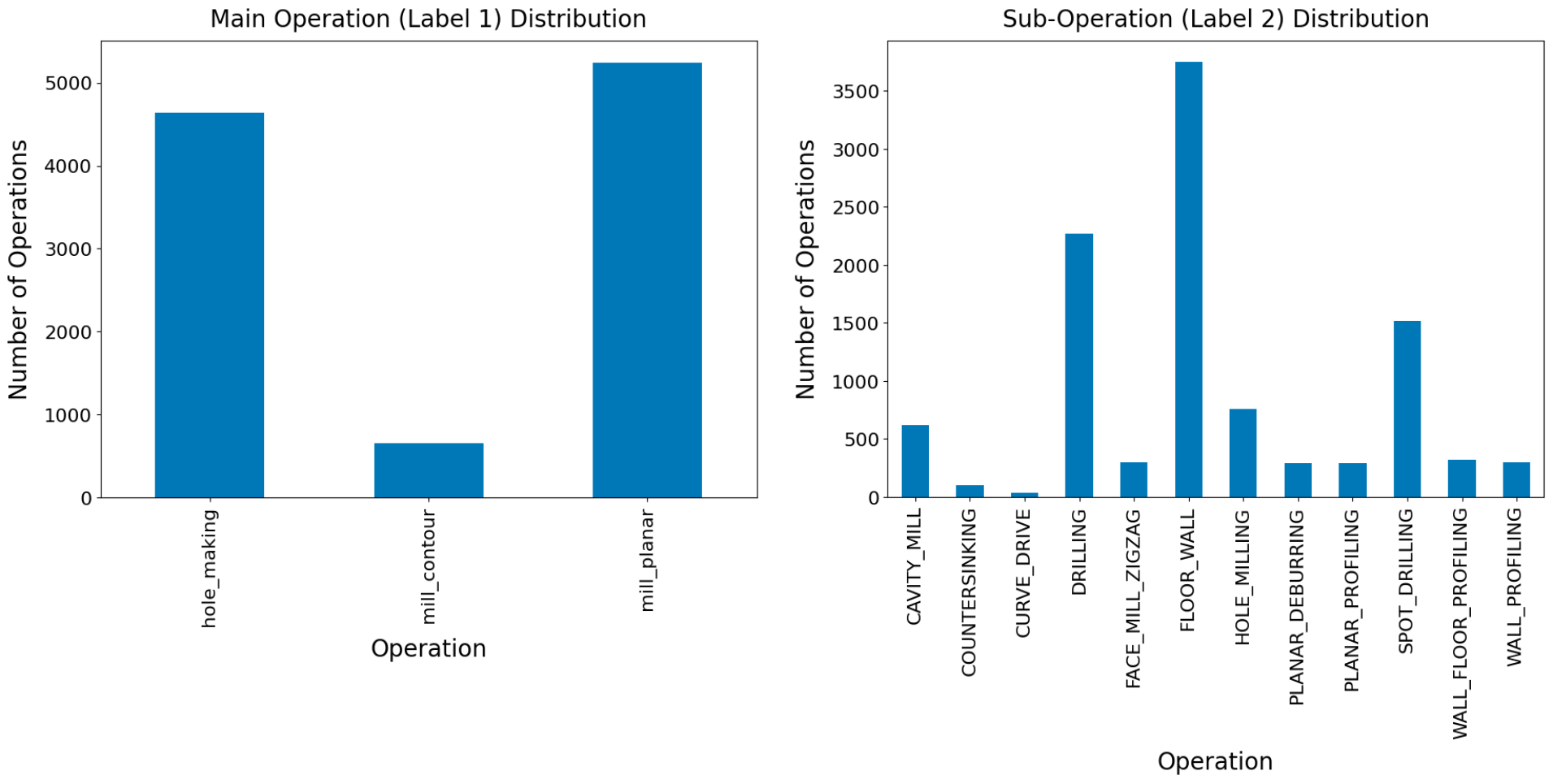}
\caption{The distribution of unique main and sub-operation labels across the dataset.
The moderate class imbalance, with dominant operation types such as hole making
accounting for the majority of instances, is an important characteristic of real-world machining operation planning.}\label{fig:labels}
\end{figure*}

The dataset comprises 2,891 instances, each consisting of a B-Rep graph representing
part design geometry and a corresponding machining operation sequence. Each sequence
contains multiple timesteps, where each timestep includes two operation labels: a
main operation type (Label 1) and a sub-operation type (Label 2). The distribution of unique labels for each operation type is shown in Figure~\ref{fig:labels}, reflecting the diversity of machining operations present across simple and complex part geometries. The label space is moderately imbalanced: a small number of main operation types (e.g., hole making and mill planar) account for the majority of
instances, while several sub-operation types appear considerably less frequently.
This imbalance is an important characteristic of the dataset, as it reflects the
natural skew of real-world machining workflows where a few fundamental operations
dominate, and it sets an expectation that Label~1 and Label~2 will present
different levels of prediction difficulty during training.

\begin{figure*}
\centering\includegraphics[width=1\linewidth]{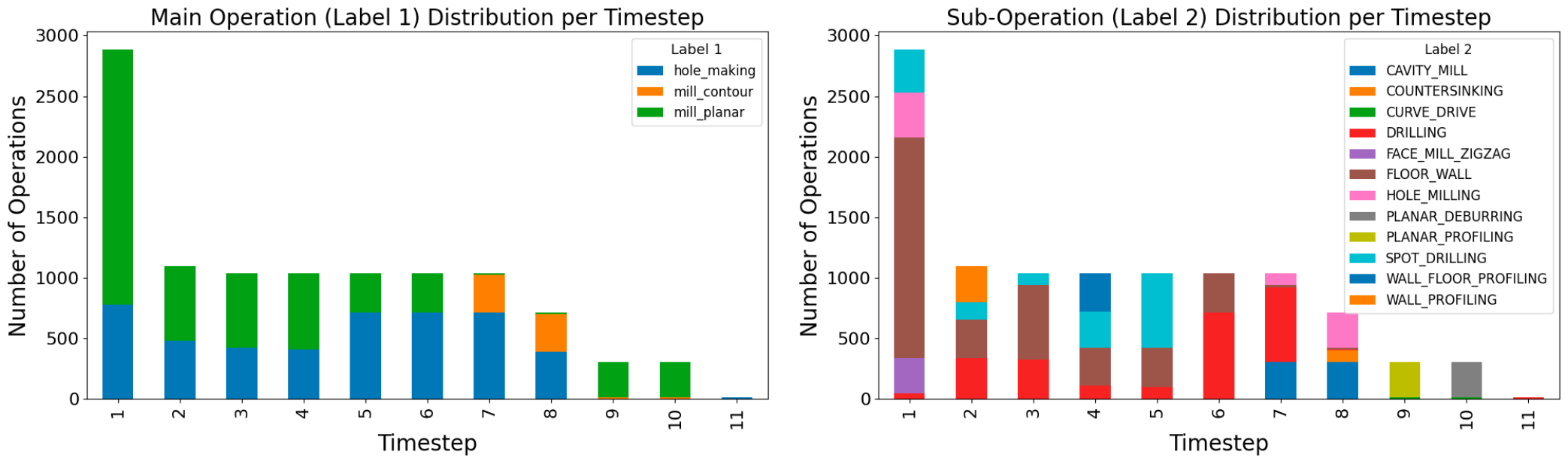}
\caption{The distribution of main and sub-operations across sequence timesteps.
The consistent ordering patterns, with coarse operations concentrated at early
timesteps and finishing operations at later ones, reflecting real-world precedence
constraints of machining operation planning.}\label{fig:labels1}
\end{figure*}

Figure~\ref{fig:labels1} presents the distribution of main and sub-operations
across sequence timesteps. This view reveals that certain operations consistently
appear at early timesteps, such as floor wall,
while planar deburring tends to occur at later positions in the
sequence. These timestep-specific patterns reflect real-world precedence constraints
in process planning, where material removal must proceed from coarse to fine before
precision operations can be applied. The consistency of these ordering patterns
across the dataset provides the transformer decoder with a learnable sequential
structure, which is a key condition for the autoregressive decoding strategy
adopted in this work.

The data was split into 80\% for training (2,393 instances) and 20\% for testing
(598 instances), with the split applied randomly while maintaining proportional
representation of both geometry classes. Hyperparameter tuning was conducted by
systematically varying the learning rate, batch size, and number of training epochs,
while keeping the model architecture fixed. Performance was evaluated using
accuracy, which measures correctly predicted labels across all timesteps and
instances for main operations (Label 1), sub-operations (Label 2), and sequence
length:

\begin{equation}
\text{Acc}^{(k)} = \frac{\sum_{i=1}^{N} \sum_{t=1}^{T_i} \mathbf{1}\left[\hat{y}_{i,t}^{(k)} = y_{i,t}^{(k)}\right]}{\sum_{i=1}^{N} T_i}, \quad k \in \{1, 2\}
\end{equation}
\begin{equation}
\text{Acc}^{\text{len}} = \frac{1}{N} \sum_{i=1}^{N} \mathbf{1}\left[\hat{T}_i = T_i\right]
\end{equation}
where $N$ is the number of test instances, $T_i$ is the sequence length of instance 
$i$, $y_{i,t}^{(k)}$ is the ground truth label, $\hat{y}_{i,t}^{(k)}$ is the 
predicted label at timestep $t$, and $\hat{T}_i$ is the predicted sequence length 
for instance $i$.

\subsection{Results}

\begin{figure}[t]
    \centering
    \includegraphics[width=0.5\textwidth]{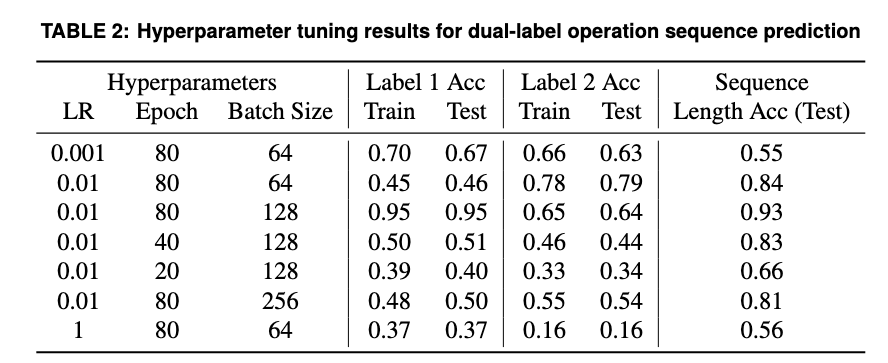}
    \label{}
\end{figure}

\begin{figure}[t]
    \centering
    \includegraphics[width=0.5\textwidth]{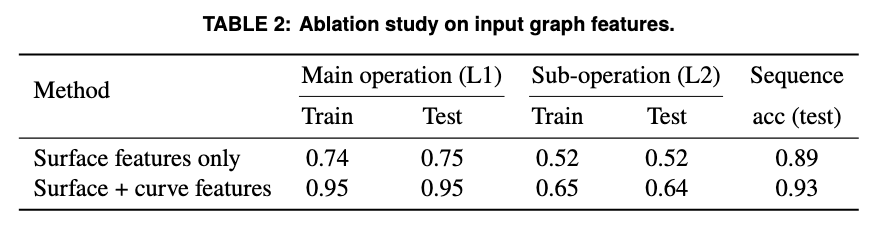}

    \label{}
\end{figure}

\begin{figure}[t]
    \centering
    \includegraphics[width=0.45\textwidth]{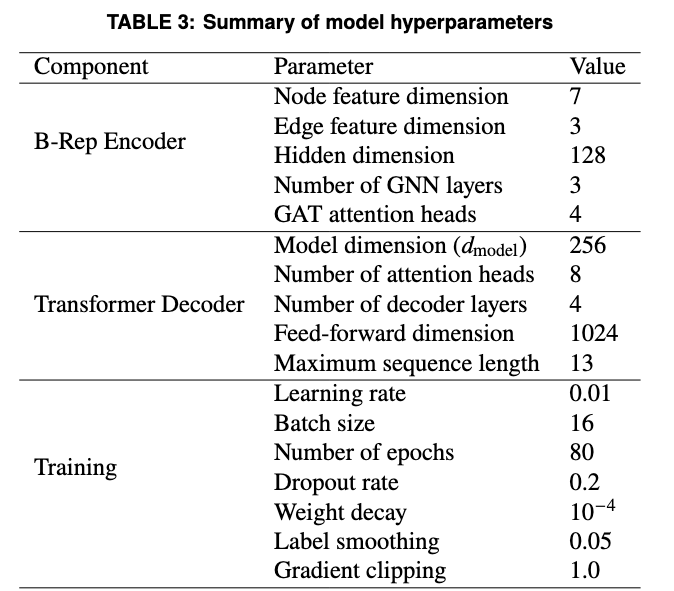}
    \label{}
\end{figure}

\begin{figure}[htbp]
    \centering
    \begin{subfigure}[b]{0.5\textwidth}
        \centering
        \includegraphics[width=\textwidth]{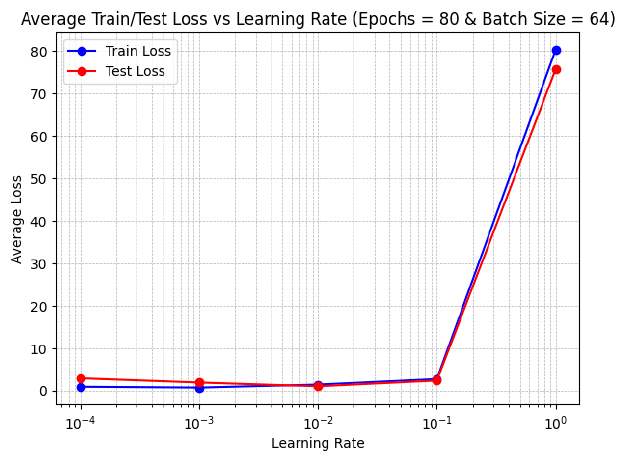}
        \caption{Epoch=80 and Batch=64}
        \label{fig:loss_a}
    \end{subfigure}
    \hfill
    \begin{subfigure}[b]{0.5\textwidth}
        \centering
        \includegraphics[width=\textwidth]{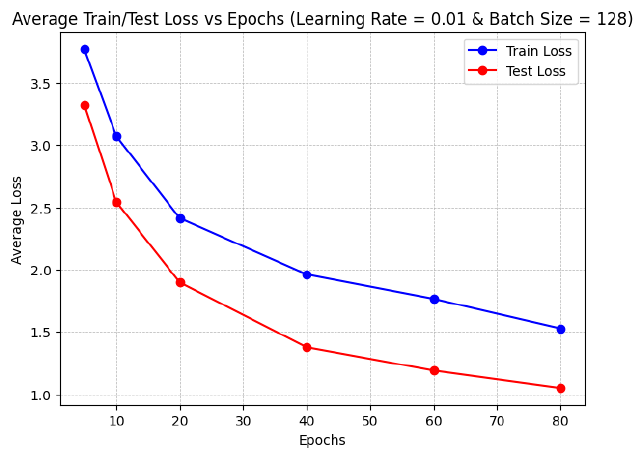}
        \caption{LR=0.01 and Batch=128}
        \label{fig:loss_b}
    \end{subfigure}
    \hfill
    \begin{subfigure}[b]{0.5\textwidth}
        \centering
        \includegraphics[width=\textwidth]{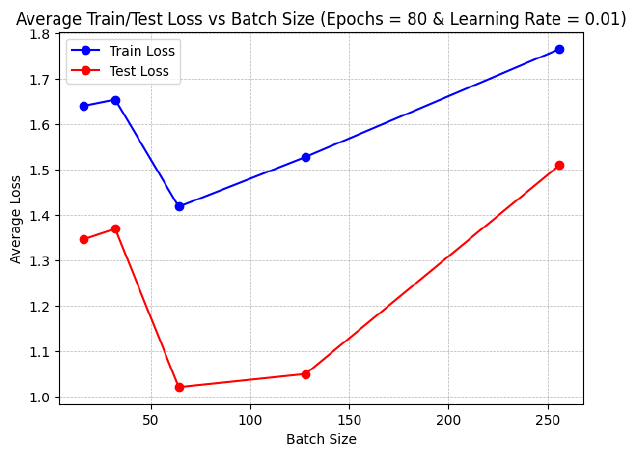}
        \caption{LR=0.01 and Epoch=80}
        \label{fig:loss_c}
    \end{subfigure}
    \caption{Training and validation loss curves for different hyperparameter configurations. The best performance is achieved with (b) LR=0.01, Epoch=80, Batch=128.}
    \label{fig:loss_curves}
\end{figure}
\begin{figure}[t]
    \centering
    \includegraphics[width=0.5\textwidth]{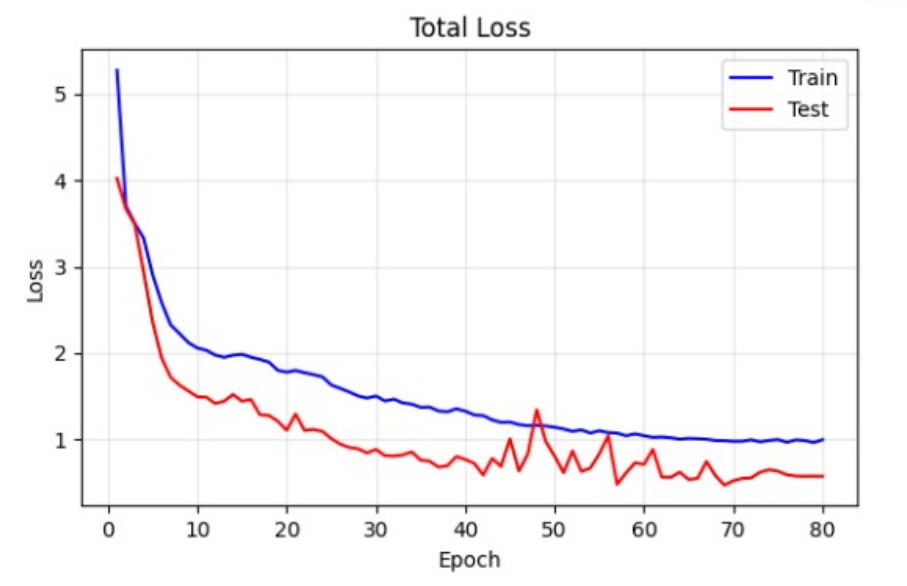}
    \caption{Loss curves of the best hyperparameter values.}
    \label{fig:loss}
\end{figure}
\begin{figure}[htbp]
    \centering
    \includegraphics[width=0.5\textwidth]{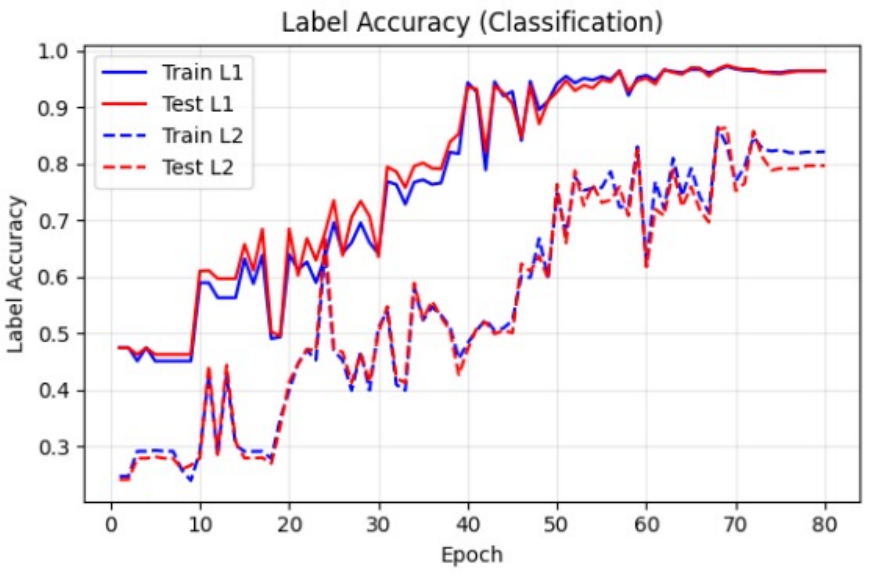}
    \caption{Training and testing accuracy curves for main operations and sub-operations.}
    \label{fig:accuracy}
\end{figure}

\begin{figure*}[!tb]
\centering\includegraphics[width=\linewidth]{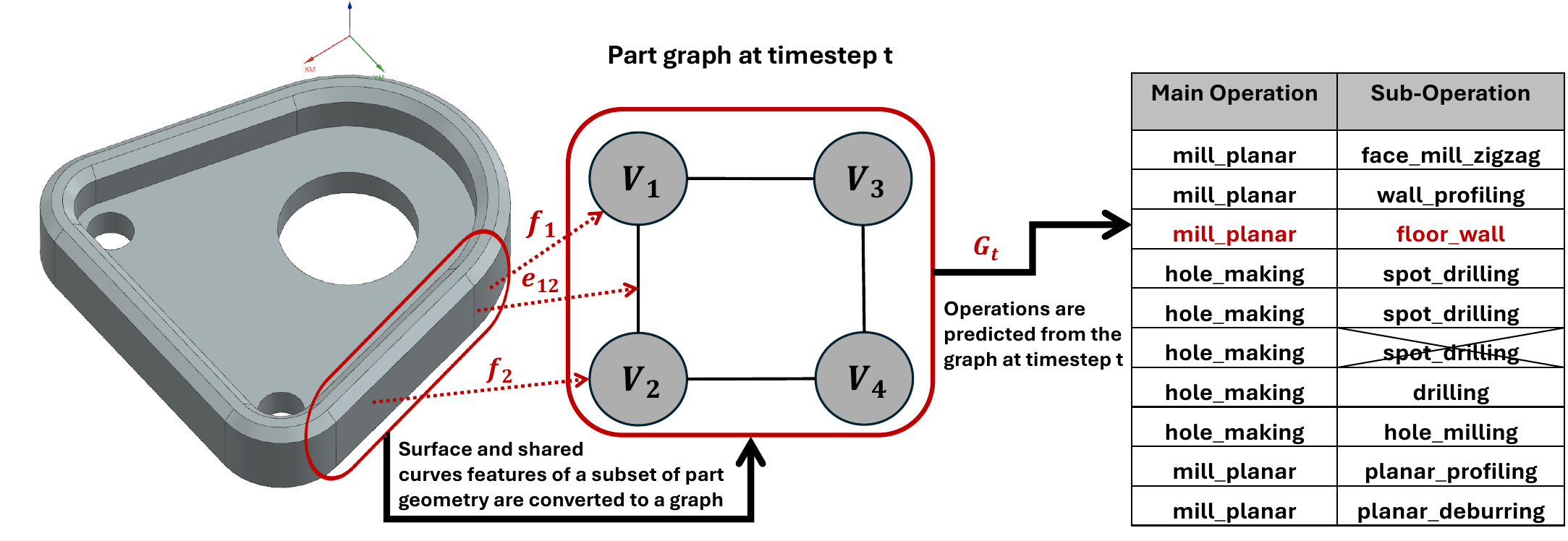}
\caption{Sample predicted machining operations for one part with 100\% and 90\% accuracies for  main operation and sub-operation, respectively.}\label{fig:Sample}
\end{figure*}

Training and testing performance across various hyperparameter settings are shown
in Table~1, and the final model configuration is
summarized in Table~3. Under the best hyperparameter configuration
(LR~=~0.01, Epoch~=~80, Batch Size~=~128), the model achieved a main operation
accuracy of 95\%, a sub-operation accuracy of 64\%, and a sequence length accuracy
of 93\% on the test set.

The hyperparameter sensitivity results in Table~1
reveal several patterns that shed light on the model's learning dynamics. Higher
learning rates (LR~=~1) led to unstable training and poor performance across all
metrics, indicating that the loss landscape for this graph-to-sequence task requires
careful gradient step control. Conversely, a lower learning rate (LR~=~0.001)
produced moderate label accuracy but significantly degraded sequence length accuracy
(55\%), suggesting that the model underfits the temporal structure of the operation
sequence when convergence is too slow. Reducing the number of training epochs (20
or 40) consistently degraded all metrics, confirming that 80 epochs are necessary
for the GNN encoder and transformer decoder to jointly converge. Increasing the
batch size beyond 128 to 256 also resulted in performance degradation, most likely
because larger batches reduce stochastic gradient noise — a regularization effect
that is particularly important for generalization on datasets of this scale.

Figure~\ref{fig:loss_curves} presents training and validation loss curves for three
representative hyperparameter configurations, enabling a direct comparison of
convergence behavior. The best configuration (LR~=~0.01, Epoch~=~80, Batch~=~128),
shown in Figure~\ref{fig:loss_curves}(b), exhibits smooth, monotonic convergence
with training and validation losses that decrease in close alignment throughout
training, indicating neither underfitting nor overfitting. By contrast,
configurations with insufficient epochs (Figure~\ref{fig:loss_curves}(c)) show
early plateauing, while those with mismatched learning rates display erratic or
stagnant loss profiles. The loss curves for the optimal configuration are further
detailed in Figure~\ref{fig:loss}, confirming that the model reaches a stable
minimum without divergence.

The accuracy curves in Figure~\ref{fig:accuracy} show training and testing
trajectories for both Label~1 and Label~2 under the best configuration. Label~1
accuracy converges faster and to a higher value than Label~2, which reflects the
underlying structure of the prediction task. Main operation types are more directly
encoded in the global topology of the B-Rep graph: a part with a single pocket
unambiguously requires a pocket milling operation, whereas predicting the precise
sub-operation variant demands finer geometric discrimination. This gap is therefore
consistent with the label distribution observed in Figure~\ref{fig:labels}, where
sub-operations exhibit greater class diversity and lower per-class frequency. Future
improvements to sub-operation prediction may benefit from richer edge feature
representations or hierarchical decoding strategies that explicitly model the
dependency between Label~1 and Label~2 at each timestep.

The 93\% sequence length accuracy is particularly significant for practical process
planning. Correctly predicting the number of manufacturing steps is a prerequisite
for generating executable operation sequences: an incorrect sequence length would
propagate errors to every downstream step. This result demonstrates that the model
reliably extracts global manufacturing complexity from B-Rep geometry, even when
fine-grained sub-operation prediction remains more challenging. The ablation study in Table~2 further clarifies the contribution
of geometric features to overall performance. Incorporating curve features alongside
surface features substantially improves both Label~1 and Label~2 accuracy, as well
as sequence length accuracy. This confirms that edge-level geometric information, 
such as curvature type, encodes manufacturing-relevant cues not
captured by surface-level attributes alone, and validates the design choice to
include both feature types in the B-Rep encoder.

A representative prediction example is shown in Figure~\ref{fig:Sample}, presenting
the complete predicted machining operation sequence for a complex part. The model
achieves 100\% accuracy on Label~1 and 90\% accuracy on Label~2 for this instance,
correctly identifying both the type and ordering of all major machining operations.
The single Label~2 error occurs at a late timestep involving a fine finishing
sub-operation, which is consistent with the general pattern observed across the
test set where sub-operation errors are concentrated at higher-index timesteps
corresponding to detail and finishing passes. This example demonstrates that the
model is capable of generating complete, coherent, and practically interpretable
operation sequences directly from B-Rep geometry as the central objective
of the proposed approach.

\subsection{Discussion}
The experimental results provide several insights into how the proposed framework connects two research streams: B-Rep representation learning and automated machining process planning. From the B-Rep representation learning perspective, the results show that topology-aware graph representations can preserve manufacturing-relevant geometric and relational information from original CAD geometry. In particular, the performance improvement obtained by incorporating both surface and curve features suggests that face-level attributes and edge-level topological relationships jointly contribute to learning process-relevant part representations. From the machining process planning perspective, the high main-operation and sequence-length prediction accuracies indicate that the learned B-Rep graph embeddings can support geometry-driven operation sequencing without requiring intermediate machining-state representations or explicit feature-recognition steps. These findings suggest that the proposed approach bridges geometric representation learning and process sequence generation, providing an upstream decision-support approach for subsequent manufacturability evaluation and process refinement.

While the proposed framework demonstrates strong performance in predicting
machining operation sequences directly from B-Rep geometry, several practical
manufacturing considerations remain outside the scope of the current study.
The current framework primarily focuses on geometry-driven sequencing and
learns precedence relationships between operations from historical process
patterns. However, real-world machining process planning often involves
additional manufacturability constraints, including tool accessibility,
fixture setup planning, machine availability, tolerance requirements, and
shop-floor specific operational preferences. These constraints are typically
determined during downstream process planning stages and are not explicitly
modeled in the current framework. Therefore, the proposed method should be
viewed as an upstream decision-support tool for automated operation sequencing
rather than a complete replacement for full computer-aided process planning systems.

Additionally, several limitations of this study should be acknowledged.
First, although synthetic data enables controlled supervision and exact
alignment between CAD geometries and operation sequences, the current dataset
does not fully capture real-world variability such as fixture constraints,
machine-specific tooling preferences, tolerance-driven planning decisions,
and noisy industrial CAD geometries. Second, the current dataset primarily
focuses on prismatic geometries, limiting generalization to freeform or highly
complex part designs. Third, although the model achieves strong performance
for primary operation prediction, sub-operation prediction remains more
challenging due to greater class diversity and lower sample frequency.
Future work will evaluate the proposed framework on industrial datasets,
incorporate richer manufacturability constraints, and improve sequence
generation for more complex manufacturing environments.

\section{Conclusion}
\label{sec:conclusion}
This paper presented a graph transformer-based framework for automated MP directly from B-Rep part geometry. The proposed framework constructs 
topology-aware graphs from B-Rep entities and generates ordered machining operation 
sequences autoregressively via transformer decoders. Experiments on 2,891 
synthetic prismatic parts demonstrated 95\% main operation accuracy, 64\% 
sub-operation accuracy, and 93\% sequence length accuracy, confirming the viability 
of B-Rep geometry-driven learning for automated process planning. Future work will focus 
on extending the dataset to more diverse and complex part geometries to provide more generalizable
operation sequence predictions.

\section*{Acknowledgments}
The authors would like to acknowledge the support of Siemens through a Senior Research Internship undertaken by the first author. The authors also appreciate the synthesized data provided by Siemens. This research was funded by NSF Grant \#2229260.

\nocite{*}

\bibliographystyle{asmeconf}  
\bibliography{asmeconf-sample}

\end{document}